\documentclass{article}
\usepackage{iclr2026_conference,times}

\usepackage{amsmath,amsfonts,bm}

\def\eqref#1{equation~\ref{#1}}
\def\1{\bm{1}}

\DeclareMathAlphabet{\mathsfit}{\encodingdefault}{\sfdefault}{m}{sl}
\SetMathAlphabet{\mathsfit}{bold}{\encodingdefault}{\sfdefault}{bx}{n}

\usepackage{hyperref}
\usepackage{url}
\usepackage{array}
\usepackage{booktabs}
\usepackage{multirow}
\usepackage{xcolor}
\usepackage{graphicx}
\usepackage{subcaption}
\usepackage{wrapfig}
\usepackage{float}
\usepackage{amsthm}

\iclrfinalcopy

\title{Teaching Agents the ``Principle", Not the ``Reward": Prescriptive Policy Learning}

\author{Shan Yang \\
National University of Singapore \\
\texttt{yang-shan@nus.edu.sg}}

\begin{document}

\maketitle

\begin{abstract}
Reinforcement learning has achieved strong performance in many tasks by letting agents learn from nothing more than a reward signal. However, in more complex tasks, agents still struggle to learn good, generalizable policies, even with carefully designed rewards and abundant samples. We argue that this difficulty stems from the \textit{reward decoding burden}: designers encode their knowledge of the desired behavior into rewards and penalties, and agents must infer this knowledge from long sequences of noisy scalars. To address this burden, we propose prescriptive policy learning (PPL), which directly teaches agents higher-level knowledge about the tasks, termed \textit{principles of behavior}, such as how to maintain balance, how to accumulate energy, and how to land smoothly. Through appropriately selected key quantities and references, PPL can directly indicate at every step whether the agent follows the principles and to what extent. Our experiments show that agents can learn good, generalizable policies without any reward. Taught only simple heuristic principles, PPL agents outperform reward-based methods with far fewer samples. Their policies also transfer directly to harder settings and keep performing well, demonstrating their strong generalization ability. In summary, PPL provides a new approach in which rough knowledge of the desired behavior is enough for agents to learn good, generalizable policies, even when the system dynamics cannot be accurately modeled.
\end{abstract}

\section{Introduction}
\label{sec:intro}

Reinforcement learning (RL) enables agents to interact directly with the environment to learn decision policies and has demonstrated strong performance in many tasks \citep{sutton2018reinforcement, mnih2015human, silver2016mastering}. However, agents typically require extensive interactions to discover desirable behavior through trial and error, which is costly or even infeasible in many real-world systems \citep{dulacarnold2021challenges}. To address this challenge, many methods incorporate prior knowledge about the task into the learning process, for example by augmenting the task reward with bonuses for progress toward the task objective or penalties for undesirable states or actions \citep{ng1999policy}. This practice is commonly known as \textit{reward design}. With effective reward design, agents in many tasks can learn desirable policies with measurably fewer interactions \citep{ng1999policy, devidze2024informativeness, zhou2024tiered}.

\begin{figure}[!htbp]
\centering
\includegraphics[width=\linewidth]{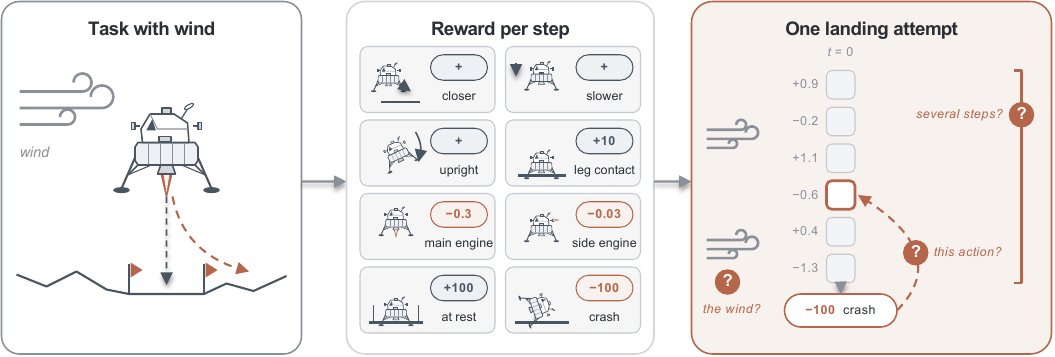}
\caption{Illustration of the reward decoding burden (Lunar Lander with wind).}
\label{fig:reward_decoding_burden}
\end{figure}

However, even with carefully designed rewards, it is still challenging for agents to learn generalizable policies. Agents must infer from an entire trajectory whether a change in the cumulative reward is caused by a particular action at a particular time or by the joint behavior over many steps, and whether it stems from the scenario or from noise. As a result, in relatively complex or noisy environments, agents require a massive number of samples to find a relatively accurate direction of policy improvement and obtain a relatively more generalizable policy \citep{cobbe2020leveraging}. Even with sufficient samples, training often leads to overfitting \citep{cobbe2019quantifying}, or agents exploit loopholes in the designed reward to obtain high returns without accomplishing the intended task \citep{pan2022effects, gao2023scaling}. We refer to this phenomenon as the \textit{reward decoding burden} and illustrate it in Figure~\ref{fig:reward_decoding_burden}.

Nevertheless, reward-based learning remains the foundation of modern RL \citep{sutton2018reinforcement}. Whether in standard deep RL algorithms \citep{mnih2015human, schulman2017proximal}, aligning large language models (LLM) with human feedback \citep{christiano2017deep}, or fine-tuning policies pretrained on demonstrations \citep{hester2018deep}, policy improvement ultimately relies on rewards. Rewards are meant to guide policy optimization, but why do they instead become an obstacle in relatively complex tasks? We argue that the core problem is that \textit{a reward compresses the knowledge about the task and the desirable behavior into a scalar}.

We take Lunar Lander as an example to illustrate this compression. The lander should land smoothly with little fuel, so designers typically penalize every engine firing and reward leg contact and approaching the landing pad. Ideally, these rewards and penalties tell agents what to do and what to avoid, and agents learn through interaction to balance different behaviors at different moments and better reach the long-term goal (e.g., a safe landing, high reward). However, because the rewards and penalties of all behaviors are compressed into reward as a scalar, agents must infer the behavioral principles from a sequence of such scalars (Figure~\ref{fig:reward_decoding_burden}). This inference often misses the ideal principles and instead overfits a mapping from the visited states to the actions that paid off there \citep{cobbe2019quantifying}. This burden is exceptionally heavy in long-horizon tasks with noisy dynamics.

In practice, humans do not learn such tasks by inferring from rewards what is good and what is bad. Instead, they learn \textit{principles of behavior} at a higher level. In obstacle avoidance, for example, one needs to adjust the speed to keep as much distance from the obstacles as possible, and in Lunar Lander, one needs to maintain the velocity and angular velocity to avoid tipping over, to approach the center, and to avoid falling to the ground too quickly. This knowledge does not take the form of a scalar. Meanwhile, directly imitating behavior from samples also learns only a mapping from states to actions rather than the principles of behavior themselves. Can we, then, teach these principles of behavior directly to agents instead of compressing them into a scalar reward?

In this work, we propose a novel algorithm named \emph{prescriptive policy learning (PPL)}, which teaches task knowledge to agents directly as principles of behavior. Such a principle is higher-level knowledge about the task. Rather than specifying concrete actions from which agents learn a mapping, it tells agents at every step how to keep the controlled object or the system in a better state. For example, a principle can give the recommended angular velocity to bring the lander upright, or the safe descent speed for the lander to land safely. This is macro-level knowledge of how the task should evolve, so it can directly guide policy optimization without relying on an accurate model of the environment. Ultimately, we anticipate that agents can learn higher-level patterns of behavior instead of struggling to decode good policies from a sequence of scalar rewards.

To the best of our knowledge, PPL is the first method in which principles of behavior alone provide the learning signal for every step of policy optimization. We evaluate PPL in three entirely different tasks, obstacle avoidance, Lunar Lander, and Reacher, which correspond to a single principle, composite principles, and multi-dimensional actions, respectively, and compare it with classic RL algorithms. Our results show that in relatively simple environments (e.g., without wind or disturbances), PPL needs significantly fewer samples than the other methods and succeeds in nearly every episode, outperforming the best of these methods in every task. When the environment becomes noisy, for example with wind, reward-based methods degrade sharply in both learning efficiency and success rate, whereas PPL shows only a modest decline in both. Importantly, policies that PPL learns in simpler settings also transfer directly to noisy settings and keep performing well. PPL therefore learns more than a mapping fitted to the training environment, namely the principles that govern the task.

\section{Prescriptive Policy Learning}
\label{sec:method}

We consider an agent that interacts with an environment in discrete steps. In each step, the agent observes a state $s\in\mathcal{S}$ and selects an action $a\in\mathcal{A}$ by a policy $\pi_\theta(a\mid s)$ with parameters $\theta$, and the environment moves to the next state $s'\in\mathcal{S}$ with the transition probability $P(s'\mid s,a)$. The agent does not know $P$ and observes it only through the transitions $(s,a,s')$ that it experiences. From a random initial state, the policy and the transition probability generate a trajectory $s_0,a_0,s_1,a_1,\ldots$ until the episode ends, and we write $\nu^{\pi_\theta}$ for the \emph{state visitation distribution}, the distribution of the states that $\pi_\theta$ visits along its trajectories.

\begin{figure}[!htbp]
\centering
\begin{subfigure}[t]{0.32\linewidth}
    \centering
    \includegraphics[width=\linewidth]{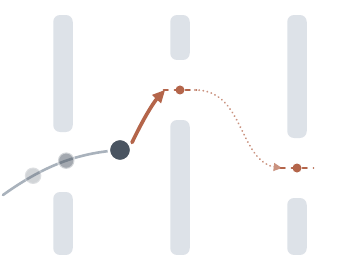}
    \caption{Obstacle Avoidance}
\end{subfigure}\hfill
\begin{subfigure}[t]{0.32\linewidth}
    \centering
    \includegraphics[width=\linewidth]{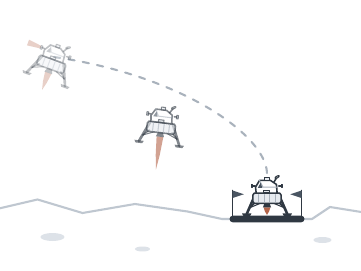}
    \caption{Lunar Lander}
\end{subfigure}\hfill
\begin{subfigure}[t]{0.32\linewidth}
    \centering
    \includegraphics[width=\linewidth]{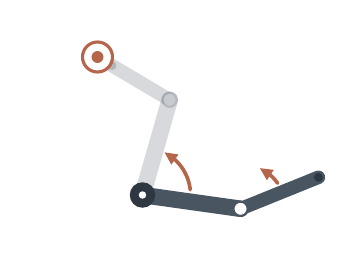}
    \caption{Reacher}
\end{subfigure}
\caption{Illustrations of different tasks.}
\label{fig:task_goals}
\end{figure}

An agent must adjust its policy to accomplish the task of its environment with a higher probability, such as passing all barriers in obstacle avoidance, coming to rest on the pad in Lunar Lander, or bringing the fingertip to the target in Reacher. In these tasks, effective decisions generally follow simple principles, such as keeping as far from the barriers as possible, staying upright above the pad without descending too fast, or bringing each joint to the angle that places the fingertip on the target \citep{ray2019benchmarking, brockman2016openai}, as shown in Figure~\ref{fig:task_goals}. How can such abstract principles of behavior be taught to an agent without relying on an accurate model of the system dynamics?

We find that we can judge whether the agent follows a principle by selecting a few key quantities $\bm{q}(s)$ of the controlled object. In obstacle avoidance, for example, the agent should stay as far from the barriers as possible when it reaches the second barrier, that is, close to the orange point in Figure~\ref{fig:task_goals}(a). We can then select the vertical velocity as the key quantity. At a given position, a vertical velocity that is too large or too small may lead to a collision, so we can design a relatively safe vertical velocity as the reference. The closer the vertical velocity is to this reference, the more safely the agent can pass the barrier. In Lunar Lander, we can likewise design an angular velocity that tilts the lander toward the pad as the reference while it moves sideways, and a descent speed that decreases as it approaches the ground as the reference while it lands. Such references need not be attainable, highly precise, or based on accurate system dynamics, and only need to point toward the goal of the principle. Designing them requires only rough knowledge of the task, such as the range of velocity changes caused by thrust, which is available in most control problems \citep{kober2013reinforcement, berkenkamp2017safe, recht2019tour, cheng2019control}. Table~\ref{tab:principle_formulations} shows how we choose the key quantity in several examples.
\begin{table}[!htbp]
\centering
\caption{Principles formulated for three tasks, with their key quantities and references.}
\label{tab:principle_formulations}
\small
\begin{tabular}{@{}
>{\raggedright\arraybackslash}p{0.12\linewidth}
>{\raggedright\arraybackslash}p{0.27\linewidth}
>{\raggedright\arraybackslash}p{0.22\linewidth}
>{\raggedright\arraybackslash}p{0.29\linewidth}
@{}}
\toprule
Task & Principle & Key quantity & Reference \\
\midrule
Obstacle avoidance
& Stay as far from the barriers as possible.
& $q=v$, the vertical velocity
& $\widetilde{v}$, the vertical velocity that a conservative deceleration brings to zero at the height of the gap center \\
\addlinespace[4pt]
Lunar Lander
& Stay upright, approach the pad, and avoid descending too fast.
& $\bm{q}=(v_y,\omega,v_x)$, the vertical, angular, and horizontal velocities
& $\widetilde{v}_y$, $\widetilde{\omega}$, and $\widetilde{v}_x$, computed from the height above the pad, the tilt, and the horizontal position and velocity \\
\addlinespace[4pt]
Reacher
& Turn each joint toward the configuration that places the fingertip on the target.
& $\bm{q}=(\dot\theta_1,\dot\theta_2)$, the joint angular velocities
& $\widetilde{q}_d$, computed from the angle that joint $d$ has left to turn \\
\bottomrule
\end{tabular}
\end{table}

A key quantity and its reference together encode a principle of behavior. In obstacle avoidance, for instance, keeping the vertical velocity close to its reference $\widetilde{q}(s)$ at every step can ensure that the agent approaches the gap center steadily. Notably, the reference does not indicate the action that the agent should take, but only provides a guideline for its behavior, which is also the core difference from knowledge provided as samples. When the environment changes, such as when the damping or the wind changes, or when the agent is at a different position and velocity, the action that it takes may be completely different. What does not change is that the agent is more likely to accomplish the task when the key quantity is closer to the reference, similar to \citet{yang2026descent}.

\begin{figure}[!htbp]
\centering
\includegraphics[width=\linewidth]{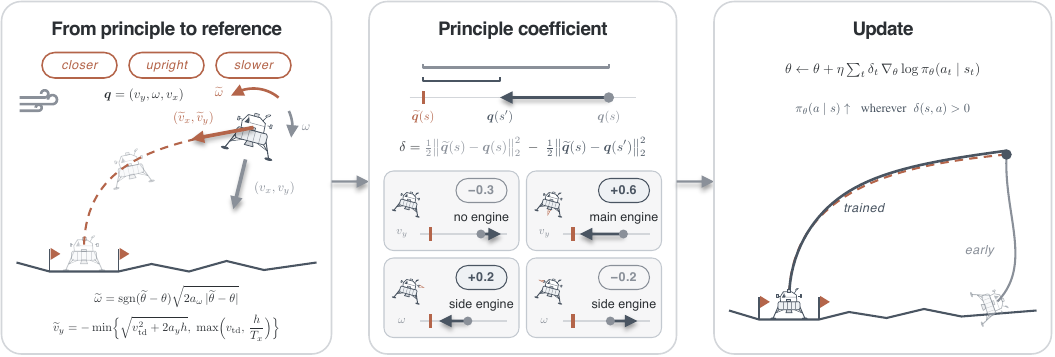}
\caption{Mechanism of PPL in Lunar Lander with wind.}
\label{fig:ppl_mechanism}
\end{figure}

We can then build a more direct learning signal based on how much each action moves the key quantity toward the reference. Here we introduce a new per-step signal, the \emph{principle coefficient} $\delta(s,a)$, which evaluates how much the transition of action $a$ in state $s$ brings the key quantity closer to its reference. Its basic form is
\begin{equation}
    \delta(s,a)
    =
    \mathbb{E}_{s'\sim P(\cdot\mid s,a)}
    \left[
        \frac{1}{2}\left\|\widetilde{\bm{q}}(s)-\bm{q}(s)\right\|_2^2
        -
        \frac{1}{2}\left\|\widetilde{\bm{q}}(s)-\bm{q}(s')\right\|_2^2
    \right],
    \label{eq:basic_coefficient}
\end{equation}
that is, the expected decrease of the deviation from the reference over one transition, with the reference fixed in state $s$ before any action is taken. The coefficient is largest for the action that brings the key quantity exactly to its reference and decreases once the key quantity overshoots it, so it does not reward larger moves beyond the reference. Then, PPL follows the update paradigm of the policy gradient \citep{williams1992simple, sutton1999policy} to update the policy parameters $\theta$,
\begin{equation}
    \theta
    \leftarrow
    \theta+\eta\,
    \mathbb{E}_{s\sim\nu^{\pi_\theta},\;a\sim\pi_\theta(\cdot\mid s)}
    \left[
        \delta(s,a)\,
        \nabla_\theta\log\pi_\theta(a\mid s)
    \right],
    \label{eq:theta_update}
\end{equation}
where $\eta>0$ is the step size and the principle coefficient of each step takes the place of the return. The update holds the visited states fixed and improves the one-step problem in each of them. The coefficient of an action is determined by its own transition, and the plan of the task enters through the reference $\widetilde{\bm{q}}(s)$. In the end, the policy raises the probability of the actions that bring the key quantity closer to its reference, and the agent learns to follow the principle at the states that it visits. Figure~\ref{fig:ppl_mechanism} illustrates this mechanism in Lunar Lander.

Although the principle coefficient is still a scalar, it carries far more information than a reward, because it can directly indicate whether the agent follows the principle and to what extent. Thus, PPL spares the agent from inferring which behavior is better through long causal chains or from noisy trajectories, and lets it learn the principle of behavior directly. Moreover, because the reference $\widetilde{\bm{q}}(s)$ gives a target and the direction in which the agent should improve its policy, instead of a mapping from states to actions, PPL can, in theory, avoid overfitting to the training environment. Overall, PPL is anticipated to learn policies that generalize better in more complex scenarios with fewer samples, which we will validate in the experiments of Section~\ref{sec:experiments}.

\subsection{Extensions to Other Task Structures}
\label{sec:method:cases}

\subsubsection{Balancing Several Principles on One Action}
\label{sec:method:multi}

When several principles act on the same action, one action may move one key quantity toward its reference and another away from it. Let the key quantities of the principles form $\bm{q}=(q_1,\dots,q_M)$ with references $\widetilde{\bm{q}}(s)$, and let each $q_i$ carry a scale $\sigma_i$, the deviation that is regarded as equally serious across key quantities. The coefficient is then Equation~\ref{eq:basic_coefficient} applied to the scaled quantities $q_i/\sigma_i$,
\begin{equation}
    \delta(s,a)
    =
    \mathbb{E}_{s'\sim P(\cdot\mid s,a)}
    \left[
        \sum_{i=1}^{M}
        \frac{1}{2}\left(\frac{\widetilde{q}_i(s)-q_i(s)}{\sigma_i}\right)^2
        -
        \frac{1}{2}\left(\frac{\widetilde{q}_i(s)-q_i(s')}{\sigma_i}\right)^2
    \right],
    \label{eq:multi_score}
\end{equation}
and PPL updates the policy with it as in Equation~\ref{eq:theta_update}. Each action receives the net change of the scaled deviations, and the scales set how a deviation in one key quantity trades off against a deviation in another. To keep the references consistent, a principle of higher priority bounds the reference of a principle of lower priority, as the upright touchdown bounds the target tilt in Lunar Lander (Appendix~\ref{app:ref:lunar}).

\subsubsection{Teaching Principles to Each Action Dimension}
\label{sec:method:multidim}

When the task can be decomposed by actuator, each dimension of the action receives its own principle and key quantity and is evaluated by the change of that key quantity. Let the action be $a=(a_1,\dots,a_D)$, let dimension $a_d$ drive the key quantity $q_d$ with reference $\widetilde{q}_d(s)$, and let $\delta_d(s,a)$ be the coefficient of Equation~\ref{eq:basic_coefficient} for $q_d$ alone. With the policy factorized as $\pi_\theta(a\mid s)=\prod_{d=1}^{D}\pi_{\theta,d}(a_d\mid s)$, PPL updates each dimension by its own coefficient,
\begin{equation}
    \theta
    \leftarrow
    \theta+\eta\,
    \mathbb{E}_{s\sim\nu^{\pi_\theta},\;a\sim\pi_\theta(\cdot\mid s)}
    \left[
        \sum_{d=1}^{D}
        \delta_d(s,a)\,
        \nabla_\theta\log\pi_{\theta,d}(a_d\mid s)
    \right].
    \label{eq:dimension_gradient}
\end{equation}
When each key quantity responds only to its own dimension, the direction of this update is the gradient of $\sum_{d}\mathbb{E}[\delta_d(s,a)]$ over the visited states. When the key quantities are coupled, the update credits the change of each key quantity to the dimension that drives it, which approximates this gradient as long as the effect of $a_d$ on the other key quantities is small.

\section{Experiments}
\label{sec:experiments}

We evaluate PPL in obstacle avoidance, Lunar Lander, and Reacher, which correspond to a single principle, composite principles, and multi-dimensional actions, respectively, and compare it with reward-based RL methods, namely proximal policy optimization (PPO) \citep{schulman2017proximal}, the deep Q-network (DQN) \citep{mnih2015human}, and soft actor-critic (SAC) \citep{haarnoja2018soft}. These methods use published hyperparameters where they exist and receive larger interaction budgets than PPL, and Appendix~\ref{app:experimental_details} gives the environments and the settings and evaluation of all methods. Taught only simple heuristic principles, PPL not only learns from far fewer samples than these methods but also keeps performing well in harder settings, including settings that it never meets in training.

\subsection{Obstacle Avoidance}
\label{sec:experiments:obstacle}

\subsubsection{Setting}
\label{sec:experiments:obstacle:setting}

An agent moves to the right at a constant speed through eight barriers and passes each barrier through a gap at a random height by thrusting up or down. Random wind perturbs its vertical velocity, and we use a moderate and a strong wind. An episode succeeds when the agent passes all eight barriers. The principle keeps the agent away from the barriers through its vertical velocity, whose reference brings the agent to rest at the height of the next gap center (Appendix~\ref{app:ref:obstacle}). The compared methods receive a reward for each barrier passed, a bonus for passing all barriers, and a penalty for a collision. PPO-R instead trains PPO with the principle coefficient of every step as its reward.

\subsubsection{Results}

\begin{figure}[!htbp]
\centering
\includegraphics[width=\linewidth]{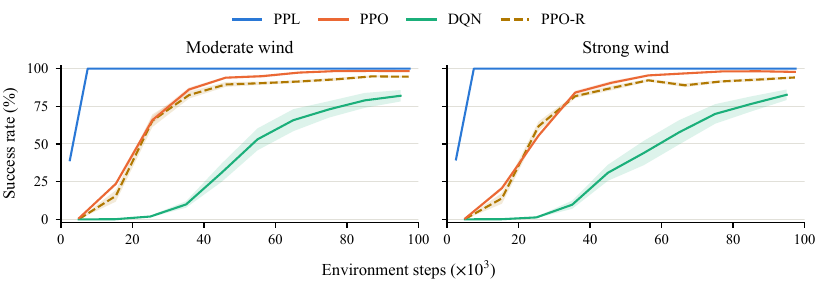}
\caption{Training success rate in obstacle avoidance.}
\label{fig:obstacle_curves}
\end{figure}

PPL learns obstacle avoidance from the fewest samples. Under both wind strengths, its success rate reaches $90\%$ after about $4\times10^3$ steps, against $10^4$ steps for PPO and about $8\times10^4$ steps for DQN (Table~\ref{tab:obstacle_results} and Figure~\ref{fig:obstacle_curves}). PPO-R needs $2\times10^4$ steps, so the same principle teaches more slowly as a reward than through PPL. After their full budgets, PPL, PPO, and PPO-R succeed in at least $96\%$ of the episodes, and DQN in $87.1\%$ and $96.6\%$.

\begin{figure}[!htbp]
\centering
\includegraphics[width=\linewidth]{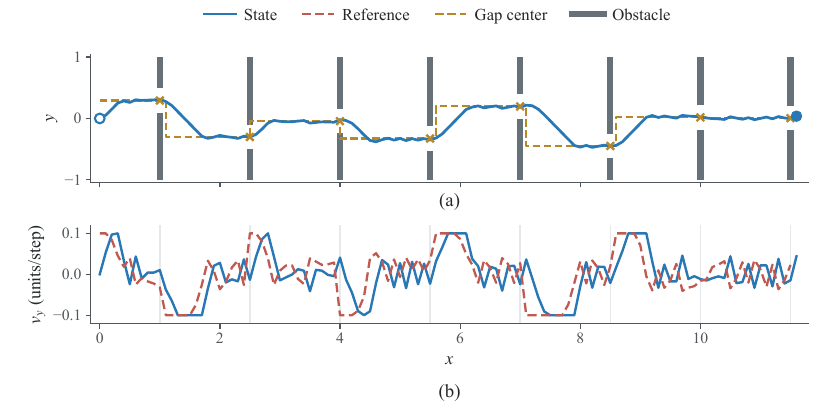}
\caption{A successful PPL episode in obstacle avoidance under moderate wind: (a) trajectory, barriers, and center of the next gap; (b) vertical velocity and its reference against horizontal position.}
\label{fig:obstacle_trajectory}
\end{figure}

The principle also shapes how the agent passes the barriers. Under moderate wind, the smallest distance of PPL from a gap edge averages $0.166$, where the half height of a gap is $0.2$, against $0.112$ for PPO, $0.097$ for DQN, and $0.053$ for PPO-R, and the ordering is the same under strong wind. The reference brings the vertical velocity to zero at the gap center, so the agent levels off at this height before each barrier (Figure~\ref{fig:obstacle_trajectory}), whereas PPO-R, rewarded by the same coefficients, passes close to the edges. PPL also thrusts in $86$ of the $116$ steps of an episode under moderate wind, fewer than PPO with $101$ and PPO-R with $115$ steps, whereas DQN thrusts least, in $71$ steps.

\begin{table}[!htbp]
\centering
\caption{Results in obstacle avoidance under moderate and strong wind.}
\label{tab:obstacle_results}
\small
\begin{tabular}{@{}lrrrrrr@{}}
\toprule
& \multicolumn{2}{c}{Success (\%)} & \multicolumn{2}{c}{Steps to $90\%$ ($\times10^3$)} & \multicolumn{2}{c}{Clearance} \\
\cmidrule(lr){2-3}\cmidrule(lr){4-5}\cmidrule(l){6-7}
Method & Moderate & Strong & Moderate & Strong & Moderate & Strong \\
\midrule
PPL & $100.0\pm0.0$ & $100.0\pm0.0$ & 4.24 & 4.18 & 0.166 & 0.155 \\
PPO & $100.0\pm0.2$ & $99.8\pm0.5$ & 10.0 & 10.0 & 0.112 & 0.107 \\
DQN & $87.1\pm24.3$ & $96.6\pm2.3$ & 75.0 & 85.0 & 0.097 & 0.087 \\
PPO-R & $99.0\pm1.4$ & $96.0\pm2.2$ & 20.0 & 20.0 & 0.053 & 0.048 \\
\bottomrule
\end{tabular}
\end{table}

Without retraining, the policies that PPL learns under moderate wind succeed in every evaluation episode under strong wind and keep passing the gaps near their centers, with an average clearance of $0.154$, close to the $0.155$ of the policies trained under strong wind. PPL therefore learns a behavior rather than a mapping from states to actions fitted to the wind seen in training.

\subsection{Lunar Lander}
\label{sec:experiments:lunar}

\subsubsection{Setting}
\label{sec:experiments:lunar:setting}

We use LunarLander-v3 \citep{brockman2016openai}, in which the lander must come to rest on a pad by firing its main engine or one of its two side engines, without and with wind. Without wind, the initial push, the terrain, and the dispersion of the engine forces are random, and the setting with wind adds wind and turbulence that vary over time. An episode succeeds when it ends with the lander at rest. The principles keep the lander upright, bring it above the pad, and slow its descent to the touchdown speed $v_{\mathrm{td}}=0.8\,\mathrm{m/s}$, with the vertical velocity $v_y$, the angular velocity $\omega$, and the horizontal velocity $v_x$ as key quantities and the references of Appendix~\ref{app:ref:lunar}. The compared methods are trained on the default reward of the environment.

\subsubsection{Results}

\begin{figure}[!htbp]
\centering
\includegraphics[width=\linewidth]{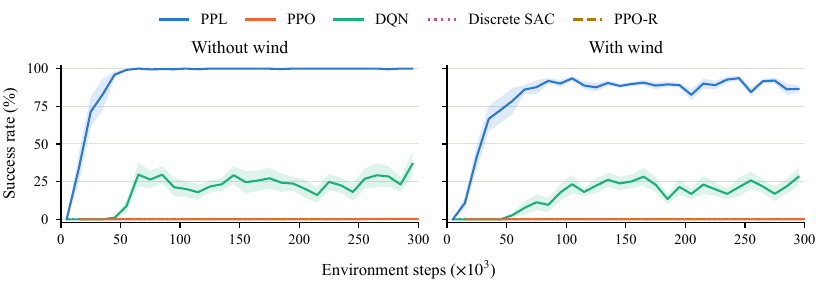}
\caption{Training success rate in Lunar Lander.}
\label{fig:lunar_curves}
\end{figure}

PPL learns to land from an order of magnitude fewer samples than PPO. Without wind, its success rate reaches $90\%$ after $2.0\times10^4$ steps, against $5.2\times10^5$ steps for PPO, and both end near full success (Table~\ref{tab:lunar_results} and Figure~\ref{fig:lunar_curves}). DQN ends at $57.2\%$ with a standard deviation of $30.5\%$ across seeds, the soft actor-critic for discrete actions (discrete SAC) \citep{christodoulou2019soft} does not learn to land, and PPO-R, trained on principle coefficients as its reward, also fails.

\begin{figure}[!htbp]
\centering
\includegraphics[width=\linewidth]{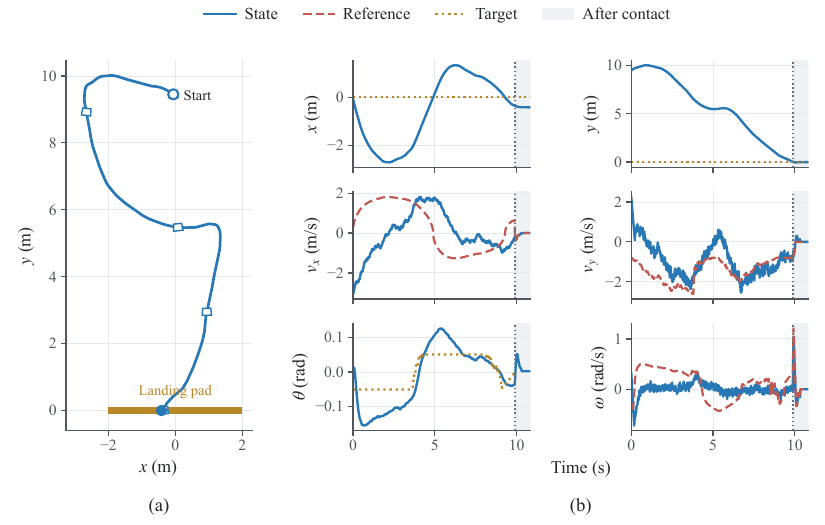}
\caption{A successful PPL landing with wind: (a) trajectory and orientations; (b) states over time, with velocity references (dashed), the pad center and the target tilt (dotted), and the period after first ground contact (shaded).}
\label{fig:lunar_trajectory}
\end{figure}

With wind, PPL remains the most reliable method. It reaches $90\%$ after $2.9\times10^4$ steps and settles at $84.6\%$, whereas PPO ends at $81.2\%$ after $10^6$ steps with a large spread across seeds, two of which never reach $90\%$. About half of the failures of PPL end at the step limit before the lander touches the ground. PPL lands at the speed that its vertical reference prescribes. Without wind, its median descent speed at the first ground contact is $0.65\,\mathrm{m/s}$, below the touchdown speed $v_{\mathrm{td}}=0.8\,\mathrm{m/s}$ of the reference, whereas PPO touches down at $0.98\,\mathrm{m/s}$, and with wind the two medians are $0.74\,\mathrm{m/s}$ and $0.56\,\mathrm{m/s}$ (Figure~\ref{fig:lunar_trajectory}).

\begin{table}[!htbp]
\centering
\caption{Results in Lunar Lander without and with wind. A dash marks a median seed that does not reach $90\%$ within its budget.}
\label{tab:lunar_results}
\small
\begin{tabular}{@{}lrrrr@{}}
\toprule
& \multicolumn{2}{c}{Success (\%)} & \multicolumn{2}{c}{Steps to $90\%$ ($\times10^3$)} \\
\cmidrule(lr){2-3}\cmidrule(l){4-5}
Method & Without wind & With wind & Without wind & With wind \\
\midrule
PPL & $100.0\pm0.2$ & $84.6\pm3.0$ & 19.7 & 29.3 \\
PPO & $98.8\pm1.5$ & $81.2\pm26.3$ & 520 & 790 \\
DQN & $57.2\pm30.5$ & $25.2\pm28.0$ & 135 & 390 \\
Discrete SAC & $0.0\pm0.0$ & $0.0\pm0.0$ & -- & -- \\
PPO-R & $0.1\pm0.1$ & $0.0\pm0.0$ & -- & -- \\
\bottomrule
\end{tabular}
\end{table}

Without retraining, the policies that PPL learns without wind land in $72.5\%$ of the episodes with wind and keep the touchdown speed of the reference, with a median descent speed of $0.63\,\mathrm{m/s}$ at the first ground contact. Most of the remaining episodes end at the step limit before the lander touches the ground. Thus, the descent that the principle teaches persists under a disturbance absent from training, and the policies lose mainly the episodes in which the lander fails to reach the ground in time.

\subsection{Reacher}
\label{sec:experiments:reacher}

\subsubsection{Setting}
\label{sec:experiments:reacher:setting}

Reacher \citep{brockman2016openai} is a planar arm with two joints that must bring its fingertip to a random target by applying a torque at each joint. We use four settings of increasing difficulty, namely the standard arm of Reacher-v5, an arm with coupled joints, an arm under gravity, and an arm under gravity with a random payload, random torque gains, and torque noise. An episode succeeds when the fingertip lies within $0.02\,\mathrm{m}$ of the target at the last step. The principle of each joint takes its angular velocity as the key quantity, with the reference of Appendix~\ref{app:ref:reacher} computed for the nominal arm in every setting, and each joint is evaluated by its own coefficient. The compared methods receive the default reward of the environment, the negative distance between the fingertip and the target minus a quadratic cost on the torques. The principle controller applies the references directly as a feedback law (Appendix~\ref{app:experimental_details}).

\subsubsection{Results}

\begin{figure}[!htbp]
\centering
\includegraphics[width=\linewidth]{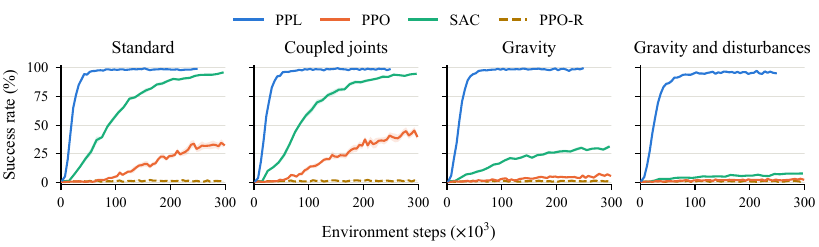}
\caption{Training success rate in the four Reacher settings.}
\label{fig:reacher_curves}
\end{figure}

PPL keeps reaching the target as the arm acquires dynamics that its principles do not model. On the standard arm and with coupled joints, PPL and SAC both succeed in about $99\%$ of the episodes, and PPL reaches $90\%$ after $4\times10^4$ steps, against $1.65\times10^5$ steps for SAC (Table~\ref{tab:reacher_results} and Figure~\ref{fig:reacher_curves}). Under gravity and under gravity with disturbances, PPL keeps $99.3\%$ and $95.2\%$, whereas SAC falls to $37.9\%$ and $9.8\%$ and PPO falls further. PPO-R stays below $2\%$ in every setting.

\begin{figure}[!htbp]
\centering
\includegraphics[width=\linewidth]{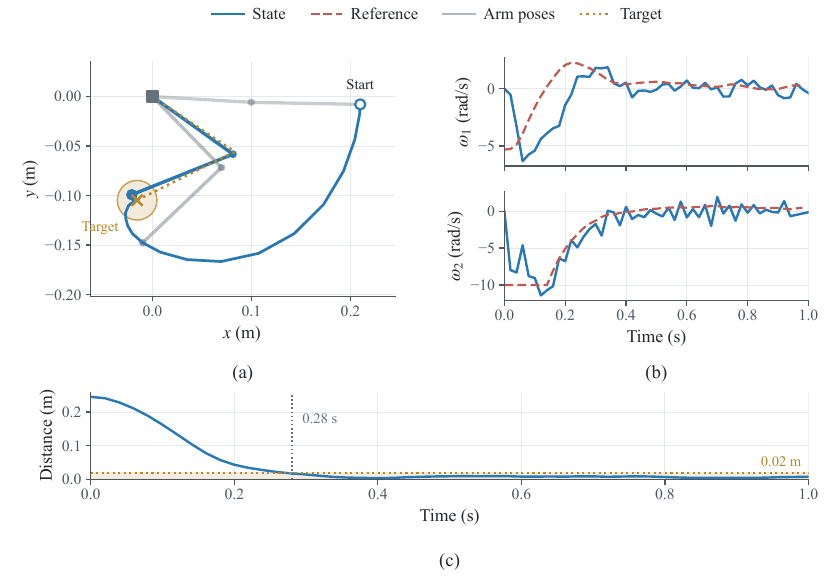}
\caption{A successful PPL episode in Reacher with gravity and disturbances: (a) fingertip trajectory, arm poses, and target configuration; (b) joint angular velocities and their references; (c) distance to the target and the $0.02\,\mathrm{m}$ success threshold.}
\label{fig:reacher_trajectory}
\end{figure}

PPL learns more than the feedback law that its references define. For suitable gains, the principle controller succeeds in all episodes on the standard arm and with coupled joints, but under gravity its best gain reaches only $69\%$, and only $59\%$ with disturbances. A torque proportional to the velocity error balances gravity only while an angle error remains, so a low gain stops short of the target and a high gain saturates the torques. Taught the same references, PPL learns from the observed transitions the torque that holds each joint at its target against gravity and the payload. In the episode of Figure~\ref{fig:reacher_trajectory}, under gravity and disturbances, the fingertip reaches the target at $0.28\,\mathrm{s}$ and stays within the success threshold until the last step.

PPL pays for holding the arm against gravity in control energy, which rises from $4.5$ on the standard arm to $17.7$ with gravity and disturbances, whereas SAC and PPO, whose reward charges the squared torque, stay below $2$.

\begin{table}[!htbp]
\centering
\caption{Final success rate (\%) in the four Reacher settings.}
\label{tab:reacher_results}
\small
\begin{tabular}{@{}lrrrr@{}}
\toprule
Method & Standard & Coupled joints & Gravity & Gravity and disturbances \\
\midrule
PPL & $99.2\pm0.6$ & $99.1\pm0.5$ & $99.3\pm0.8$ & $95.2\pm2.0$ \\
PPO & $68.2\pm14.2$ & $75.9\pm9.8$ & $14.7\pm7.4$ & $3.1\pm1.2$ \\
SAC & $98.7\pm1.2$ & $98.5\pm1.3$ & $37.9\pm5.1$ & $9.8\pm3.4$ \\
PPO-R & $0.9\pm1.5$ & $1.4\pm1.7$ & $0.6\pm0.8$ & $1.9\pm1.4$ \\
Principle controller, best gain & $100.0$ & $100.0$ & $69.0$ & $59.0$ \\
\bottomrule
\end{tabular}
\end{table}

Without retraining, the policies that PPL learns on the standard arm succeed in $98.3\%$ of the episodes with coupled joints, and those learned under gravity succeed in $86.2\%$ of the episodes with disturbances added. In both cases, the fingertip ends on average within the success threshold of $0.02\,\mathrm{m}$, at $0.005\,\mathrm{m}$ and $0.015\,\mathrm{m}$ from the target. Thus, the joint motion that the principles teach persists under dynamics absent from training.

\section{Conclusion}
\label{sec:conclusion}

In this work, we argued that a central obstacle of RL is the \textit{reward decoding burden}, since a reward compresses the knowledge about the task and the desirable behavior into a scalar that agents must decode. To address this burden, we proposed PPL, which teaches this knowledge to agents directly as principles of behavior and evaluates every action by the change that its own transition makes in a few key quantities relative to their references. In tasks with a single principle, composite principles, and multi-dimensional actions, PPL learned from far fewer samples than reward-based methods, remained reliable as disturbances made the tasks harder, and transferred to harder settings without retraining. These findings confirm that task knowledge does not have to be compressed into a reward to guide learning, and that teaching it as principles gives agents a direct signal at every step and policies that hold beyond the training environment.

PPL may give the training of agents for specific tasks more flexibility and efficiency. For embodied agents, rewards are hard to write, whereas people can describe how a behavior unfolds, much as a coach tells an athlete how to perform a movement. We also tried other control tasks, such as Mountain Car and Acrobot, in which an LLM can suggest suitable principles of behavior from its own analysis of the task. On the other hand, PPL teaches agents the behavior itself rather than relying on imitation from large numbers of demonstrations, so its policies can in theory keep working when the environment changes. Its high learning efficiency means that a new platform needs only brief retraining on the same principles, without collecting new demonstrations. Finally, learning from principles makes it easier to trace a failure back to the principle involved and to provide guarantees on safety requirements.

We see three important directions for PPL. First, hierarchical principles or stage-wise PPL can be designed to support tasks with several stages and competing goals. Second, for tasks of the same kind with different requirements, such as a gentler or a faster landing, the references can be recomputed from each requirement while the principles stay the same. Finally, by building a library of principles that can be reused across tasks, PPL can bring the same behaviors to new robots and more complex tasks with little additional design.

\bibliography{iclr2026_conference}
\bibliographystyle{iclr2026_conference}
\newpage

\appendix

\section{References of Key Quantities}
\label{app:reference_computation}

As introduced in Section~\ref{sec:method}, the key quantity is a value that the action directly affects, such as a velocity or an angular velocity, and its reference shows whether the agent follows the principle. We next describe how general knowledge of the environment and the task yields these references, so that PPL can evaluate each action from its own transition.

\subsection{Obstacle Avoidance}
\label{app:ref:obstacle}

To stay far from the barriers, the agent should pass each barrier near the center of its gap, and its thrust directly changes the vertical velocity $v$. At each height, the agent should therefore rise or sink fast when far from the height $y_{\mathrm{gap}}$ of the next gap center and slowly when close to it. We express this behavior as the speed from which a deceleration $a$ comes to rest at this height, within the speed limit $v_{\max}$ of the environment,
\begin{equation}
    \widetilde{v}
    =
    \operatorname{sgn}\left(y_{\mathrm{gap}}-y\right)
    \min\left\{v_{\max},\;\sqrt{2a\left|y_{\mathrm{gap}}-y\right|}\right\},
    \label{eq:obstacle_reference}
\end{equation}
where $v_{\max}=0.1$ and $a=0.025$ is half of the velocity change of one thrust step. An agent that rises faster than this reference near the gap center is then evaluated in favor of no thrust or downward thrust, which slow it down before it overshoots.

\subsection{Lunar Lander}
\label{app:ref:lunar}

The lander should stay upright, approach the pad, and avoid descending too fast, and two references carry these behaviors, one for the attitude and one for the descent. The side engines directly change the angular velocity $\omega$. Because the lander moves sideways only when it is tilted, it should tilt only as far as the remaining horizontal velocity needs and return to upright before touchdown. From the current tilt $\theta$, the reference therefore turns the lander toward this tilt $\widetilde{\theta}$, fast when far from it and slowly when close to it,
\begin{equation}
    \widetilde{\omega}
    =
    \operatorname{sgn}\left(\widetilde{\theta}-\theta\right)\sqrt{2a_\omega\left|\widetilde{\theta}-\theta\right|}.
    \label{eq:lunar_angular_reference}
\end{equation}
The main engine directly changes the vertical velocity $v_y$. From the current position, the lander should slow down near the ground and should not arrive at the ground before it is above the pad, so the reference takes the slower of these two descents,
\begin{equation}
    \widetilde{v}_y
    =
    -\min\left\{
        \sqrt{v_{\mathrm{td}}^2+2a_yh},\;
        \max\left(v_{\mathrm{td}},\,\frac{h}{T_x}\right)
    \right\},
    \label{eq:lunar_vertical_reference}
\end{equation}
where $h$ is the height above the pad, $v_{\mathrm{td}}$ is the speed at touchdown, $T_x$ is the time that the approach to the pad still needs, and $a_\omega$ and $a_y$ are planned decelerations. The horizontal velocity also has a reference, which grows with the distance from the pad and vanishes above it. Once a leg touches the ground, the ground brings the lander to rest, so each reference is an estimate of the next velocity with the engines off, and an action is evaluated in favor of leaving the engines off.

These references need only general knowledge of the lander. The settings below use the acceleration $a_m$ of the main engine, gravity $g$, and the angular acceleration $\alpha_s$ and horizontal acceleration $a_s$ of a side engine. To leave the engines spare capability for catching up with the references, we assume that each engine fires in half of the steps and hold back half of the deceleration that this provides, which gives $a_y=(a_m/2-g)/2$ for the upright lander, $a_\omega=\alpha_s/4$, and $a_x=a_s/4$ for the horizontal approach. To keep the lander nearly upright during the approach, we limit the tilt to $\theta_l=\arcsin(2a_x/a_m)$, at which the main engine pushes it sideways at $2a_x$. To have it upright at touchdown, we start returning the tilt to zero at twice the rotation time $2\sqrt{\theta_l/a_\omega}$. To avoid a crash at contact, we set the touchdown speed to $v_{\mathrm{td}}=0.8\,\mathrm{m/s}$, below the largest speed that the lander survives. The scales treat deviations of $1\,\mathrm{m/s}$ in $v_y$, $0.3\,\mathrm{rad/s}$ in $\omega$, and $0.5\,\mathrm{m/s}$ in the horizontal velocity as equally serious.

\subsection{Reacher}
\label{app:ref:reacher}

To bring the fingertip to the target, each joint should turn toward the configuration that places the fingertip there and slow down on arrival, and its torque directly changes its angular velocity $q_d=\dot\theta_d$. The link lengths give the angle $\theta_d^\star$ of each joint in this configuration, and of the two configurations that reach the target, we take the one closer to the current configuration. The joint should turn fast when it has far to go and slowly as it arrives, so the reference is proportional to the angle left to turn,
\begin{equation}
    \widetilde{q}_d
    =
    \operatorname{clip}\left(k\left(\theta_d^\star-\theta_d\right),\;-q_{\max},\;q_{\max}\right),
    \label{eq:reacher_reference}
\end{equation}
where the angle difference of the shoulder is wrapped to $[-\pi,\pi]$, $k=10\,\mathrm{s}^{-1}$, and $q_{\max}=10\,\mathrm{rad/s}$. Each torque is then evaluated by how far it moves the angular velocity of its own joint toward this reference, and the agent learns from the transitions which torque achieves this.

\section{Experimental Details}
\label{app:experimental_details}

\subsection{Environments}
\label{app:environments}

Obstacle avoidance is our own environment. The agent moves to the right by $0.1$ per step through eight barriers spaced $1.5$ apart, each with a gap of height $0.4$ at a random height, and its thrust $u_t\in\{-1,0,1\}$ and the wind $w_t$ change its vertical velocity $v_t$ and height $y_t$ as
\begin{equation}
    v_{t+1}
    =
    \operatorname{clip}\left(0.8\,v_t+0.05\,u_t+w_t,\;-0.1,\;0.1\right),
    \qquad
    y_{t+1}
    =
    \operatorname{clip}\left(y_t+v_{t+1},\;-1,\;1\right),
    \label{eq:obstacle_dynamics}
\end{equation}
where $w_t$ is Gaussian with standard deviation $w_{\max}/2$, clipped to $[-w_{\max},w_{\max}]$, with $w_{\max}=0.015$ for the moderate wind and $w_{\max}=0.03$ for the strong wind. The observation contains $y_t$, $v_t$, and the horizontal distance and vertical offset to the next two gaps. The compared methods receive $10$ for each barrier passed, a further $100$ for passing all barriers, and $-10$ for a collision.

Lunar Lander and Reacher are standard environments, and we state only what we change. The setting with wind in Lunar Lander uses wind power $5$ and turbulence power $1.5$. In Reacher, the coupled joints use armature $2\times10^{-4}$, damping $10^{-3}$, and gear $0.15$, gravity acts in the plane of the arm, and the disturbances add a payload at the fingertip drawn from $[0,0.02]\,\mathrm{kg}$ and a torque gain for each joint drawn from $[0.7,1.3]$ in every episode, together with Gaussian torque noise of standard deviation $0.1$ in every step.

\subsection{Training settings of PPL}

We compute the coefficient of Equation~\ref{eq:multi_score} from each sampled transition in a batch, subtract a baseline fitted on the state over the batch, and normalize the result by its standard deviation over the batch. We reuse each batch for several epochs of minibatch updates, because the coefficient of a transition is determined by the transition alone and remains informative after the policy changes, and we clip the likelihood ratio between the current policy and the policy that collected the batch \citep{schulman2017proximal}. In Reacher, each joint has its own coefficient, baseline, normalization, and ratio, and the baseline of a joint also takes the torque of the other joint as input, which leaves the expected direction unchanged because the torques are sampled independently. All networks use two hidden layers of 64 units with hyperbolic tangent activations, and the baseline is fitted by 20 full-batch steps of Adam with learning rate $10^{-3}$ in every batch.

\begin{table}[!htbp]
\centering
\caption{Training settings of PPL.}
\label{tab:ppl_hyperparameters}
\small
\begin{tabular}{@{}llll@{}}
\toprule
Setting & Obstacle avoidance & Lunar Lander & Reacher \\
\midrule
Policy & Categorical, 3 actions & Categorical, 4 actions & Gaussian, 2 torques \\
Policy input & Raw observation & Raw observation & Normalized observation \\
Baseline input & State & State & State, other torque, error \\
Trajectories per batch & $4$ & $4$ & $20$ \\
Epochs per batch & $30$ & $30$ & $10$ \\
Minibatch size & $256$ & $256$ & $64$ \\
Clip range & $0.2$ & $0.2$ & $0.2$ \\
Learning rate & $3\times10^{-4}$ & $3\times10^{-4}$ & $3\times10^{-4}$ \\
Gradient-norm clip & $0.5$ & $0.5$ & $0.5$ \\
Budget (steps) & $2\times10^5$ & $3\times10^5$ & $2.5\times10^5$ \\
Evaluation interval (steps) & $4000$ & $2000$ & $10^4$ \\
\bottomrule
\end{tabular}
\end{table}

\subsection{Compared methods}

We run DQN and SAC, and PPO except in Reacher, with Stable-Baselines3 \citep{raffin2021stable}. In obstacle avoidance, PPO and DQN use its default hyperparameters with a budget of $5\times10^5$ steps. In Lunar Lander, PPO and DQN use the hyperparameters of RL Baselines3 Zoo for LunarLander-v3, with budgets of $10^6$ and $5\times10^5$ steps. In Reacher, SAC uses the default hyperparameters with $10^4$ initial random steps and a budget of $3\times10^5$ steps, and PPO uses our implementation with a budget of $10^6$ steps, 40 trajectories per batch, 10 epochs, minibatches of 64, clip range $0.2$, learning rate $3\times10^{-4}$, discount factor $0.99$, and generalized advantage estimation parameter $0.95$. PPO-R uses the settings of PPO in each task, and in Lunar Lander it takes the current velocities as its ground references. The principle controller applies $a_d=\operatorname{clip}(k_p(\widetilde{q}_d-\dot\theta_d),-1,1)$ with the gains $k_p\in\{0.05,0.1,0.25,0.5,1,2,4\}$, and Table~\ref{tab:reacher_results} reports the best of them. PPO and DQN are evaluated every $10^4$ steps in obstacle avoidance and Reacher, and every $2\times10^4$ and $10^4$ steps in Lunar Lander.

\subsection{Evaluation protocol}

Every method is trained with seeds $0$ to $9$. Intermediate evaluations use $50$ episodes in obstacle avoidance and Lunar Lander and $100$ in Reacher, and final evaluations use $200$ and $100$, with environment seeds from $10000$ that fix all random perturbations, so all methods face the same episodes. The policy takes its deterministic action in every evaluation, and transfer evaluates the final policies of one setting on the episodes of a harder setting without retraining. Tables report the mean $\pm$ standard deviation over seeds, and training curves show the mean with one standard error. Steps to $90\%$ is the median over seeds of the first step at which the success rate reaches $90\%$, clearance is the mean smallest distance from a gap edge in an episode, and control energy is $\sum_t\|a_t\|_2^2$ over the clipped actions. The training curves of DQN include its exploratory actions.

\end{document}